\documentclass{article}

\usepackage[preprint]{neurips_2026}

\usepackage[utf8]{inputenc} 
\usepackage[T1]{fontenc}    
\usepackage{hyperref}       
\usepackage{url}            
\usepackage{booktabs}       
\usepackage{amsfonts}       
\usepackage{nicefrac}       
\usepackage{microtype}      
\usepackage{xcolor}         
\usepackage[pdftex]{graphicx}
\usepackage{algorithm}
\usepackage{algorithmic}
\usepackage{listings}
\usepackage{amsmath}
\usepackage{stmaryrd}
\usepackage{amssymb}
\title{From 0-Order Selection to 2-Order Judgment: Combinatorial Hardening Exposes Compositional Failures in Frontier LLMs}

\author{%
  Hanmeng Liu$^{1}$, Shichao Weng$^{2}$, Xiulai Li$^{1}$, Zhicai Zhang$^{1}$, Anli Yan$^{1}$, Xiaozhang Liu$^{1}$ \\[2pt]
  $^{1}$Hainan University, Haikou, China \quad $^{2}$Fudan University, Shanghai, China \\[2pt]
  \texttt{\{liuhanmeng,lixiulai01,zzcai,yananli,lxzh\}@hainanu.edu.cn}, \texttt{scweng23@m.fudan.edu.cn}
}

\begin{document}

\maketitle

\begin{abstract}

Multiple-choice reasoning benchmarks face dual challenges: rapid saturation from advancing models and data contamination that undermines static evaluations. Ad-hoc hardening methods (paraphrasing, perturbation) attempt to increase difficulty but sacrifice logical validity for surface complexity, falling short to challenge advanced reasoning models.
We present LogiHard, a formal framework that deterministically transforms 0-order selection into 2-order logical judgment, which significantly increases the thinking overhead and reasoning steps.
The framework integrates Item Response Theory (IRT) for computerized adaptive testing (CAT), enabling precise difficulty control with fewer questions than static benchmarks.
We instantiate LogiHard-2k, a logical reasoning dataset constructed by cognitively ranking high-stakes examination questions via 9-dimensional analysis of model thinking traces, followed by combinatorial transformation of high-difficulty items. Evaluation across twelve state-of-the-art models
reveals an accuracy degradation ranging from $31\%$ to $56\%$ on combinatorially hardened questions. LLMs suffer from the multi-select failure and early exit bias, which are not shared by human testees. 
Zero-shot transfer to MMLU demonstrates $47\%$ accuracy degradation ($89.84\% \to 42.86\%$), confirming applicability across domains with provable validity preservation.
The consistent aggregate degeneration is domain-agnostic and stems not from knowledge deficits but from a combinatorial reasoning gap, reflecting a training-induced completeness-verification deficit.

\end{abstract}

\section{Introduction}

Multiple-choice questions (MCQs) remain the dominant paradigm for evaluating large language models (LLMs) \citep{mmlu2021,bbh2022,phan2025lastexam}, with logical reasoning benchmarks garnering particular attention for their ability to isolate pure reasoning from domain-specific knowledge. The recent emergence of Large Reasoning Models (LRMs) \citep{deepseekai2025deepseekr1incentivizingreasoningcapability,qwq32b} has accelerated this trend by leveraging test-time scaling and extended chain-of-thought (CoT) with reflection \citep{chen2025reasoningerasurveylong} to achieve unprecedented performance on complex reasoning tasks. 
Supervised fine-tuning on logical reasoning datasets like LogiQA \citep{10174688,liu2023logicot} has become common practice for instilling foundational reasoning capabilities \citep{muennighoff2025s1simpletesttimescaling,nvidia2025nvidianemotronnano2}.

Yet the development of LRMs has led to the rapid saturation of reasoning benchmarks. MMLU falls to GPT-5 at 92.5\% \citep{singh2026openaigpt5card}, Sonnet 3.5 exceeds 93.1\% on BBH \citep{bbh2022}, and OpenAI o1 achieves 90.0\% average accuracy on LogiQA \citep{latif2025comparative}. These numbers signal not the resolution of machine reasoning, but the failure of static evaluation \citep{malek2025frontierllmsstrugglesimple}. 
Contemporary models achieve superhuman accuracy partly through training set memorization and exploitation of surface patterns (position bias, lexical overlap, stylistic cues)~\cite{xie-etal-2025-memorization}. 
In response, ad-hoc hardening methods have proliferated: None-of-the-Above (NOTA) distractors \citep{tam2025none,madhusudhan2025llms}, adversarial perturbations \citep{wallace2021universaladversarialtriggersattacking,moffett-dhingra-2025-close}, and template-based obfuscation \citep{gsm-symbolic}. 
These approaches, however, suffer from a fundamental \emph{validity crisis} where human verification is required for a rigorous procedure \citep{park2025vlmcont}. Stochastic perturbations introduce semantic drifting and unintended artifacts in generated questions \citep{sun2025emperor,chen-etal-2025-benchmarking-large} rather than eliminating contamination; while NOTA variants increase surface verification burden by requiring models to reject all distractors, they remain \emph{0-order selection tasks} that do not elevate logical order---the core reasoning step is still to identify a single correct option among candidates, not to evaluate compound propositional constraints. 
Consequently, they fail to challenge state-of-the-art reasoning models that already excel at extended chain-of-thought verification.

\begin{figure}
    \centering
    \caption{An example from the LogiHard-2k benchmark}
    \includegraphics[width=\linewidth]{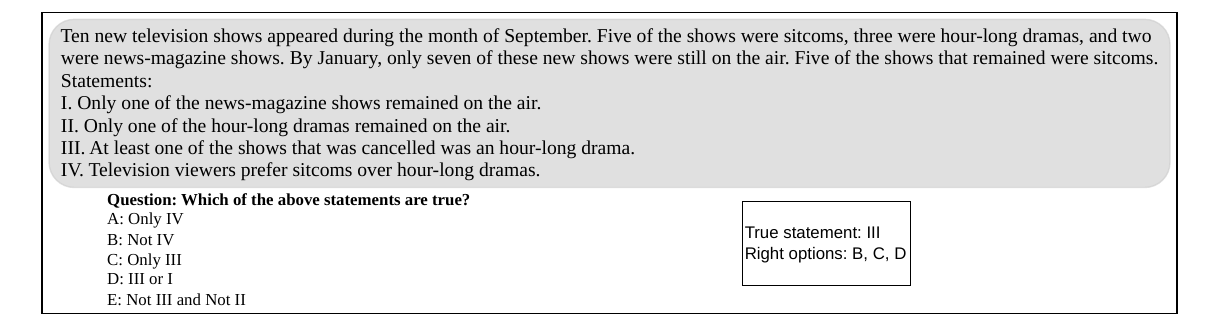}
     \vspace{-2em}
    \label{fig:example}
\end{figure}

To address this, we propose LogiHard, a formal framework for dynamic hardness evaluation that deterministically transforms multiple-choice questions via propositional logical combinatorics. 
Our core premise is that if a model truly understands logical entailment, it should handle the same knowledge when expressed as combinatorial constraints over atomic propositions. As shown in Figure~\ref{fig:example}, rather than corrupting surface text, LogiHard elevates the reasoning order by mapping atomic options to propositional variables and synthesizing compound formulas---including exactness, disjunction, and negation---that the model must evaluate under the ground-truth assignment. 
This transforms the task from 0-order selection to 2-order logical judgment, ensuring \emph{validity-by-construction} while rendering memorization ineffective.
LogiHard comprises three synergistic components:
1) A \emph{deterministic combinatorial protocol} that synthesizes logically valid, difficulty-calibrated questions via controlled propositional complexity (e.g., Easy: $\land$, Medium: $\lor$, Hard: $\neg$, Expert: compound $\neg$);
2) A \emph{cognitive difficulty scoring} module that extracts 9-dimensional metrics (oscillation points, logic density, abductive depth) from model thinking traces to empirically rank items before hardening;
3) A \emph{dynamic evaluation protocol} based on Item Response Theory (IRT) \citep{lord1980applications} and Computerized Adaptive Testing (CAT), enabling efficient ability estimation $\hat{\theta}$ with fewer questions than static benchmarks.

To rigorously validate this framework, we instantiate \textbf{LogiHard-2k}, a native logical reasoning dataset curated from high-stakes human examinations, isolating pure logical deduction via syllogistic, analogical, and propositional reasoning. 
Evaluation across twelve state-of-the-art models reveals a fundamental combinatorial multi-select reasoning failure---all systems exhibit severe degradation despite strong conventional benchmark performance---confirming that current models lack robust propositional constraint satisfaction.

Our contributions are threefold:
\begin{enumerate}
    \item \textbf{Validity-guaranteed hardness synthesis.} A deterministic combinatorial protocol that transforms MCQs into propositional logic tasks via validity-by-construction, elevating reasoning from 0-order selection to 2-order judgment while resisting memorization---the correct answer determines the truth assignment, yet the atomic option never appears directly in the output, rendering memorization ineffective while eliminating the semantic drift of adversarial methods.
    \item \textbf{Dynamic difficulty control.} We establish adaptive evaluation via two core mechanisms: (i) \emph{on-set} difficulty control through the combinatorial protocol's tiered operator configurations, enabling deterministic calibration of item complexity at generation time; and (ii) IRT-based computerized adaptive testing (CAT), where item parameters are informed by automated 9-dimensional cognitive scoring, enabling precise ability targeting and efficient testing.
    \item \textbf{Identifying reasoning failure and cross-domain transfer.} We instantiate LogiHard-2k and evaluate \textbf{twelve} state-of-the-art models, revealing a multi-select failure and combinatorial reasoning gap (accuracy degradation $31$--$56\%$) with zero logical invalidity. Zero-shot transfer to MMLU confirms cross-domain generalization ($89.84\% \to 42.86\%$).
\end{enumerate}

\section{Related Work}

Our work builds upon and extends three critical research strands: systematic methods for hardening multiple-choice benchmarks, combating data contamination and benchmark saturation, and adaptive evaluation paradigms. While existing approaches address isolated aspects of these challenges, LogiHard provides a unified framework that combines \textit{validity-guaranteed hardness synthesis} with \textit{dynamic difficulty adaptation}.

\subsection{Benchmark Hardening and Surface Perturbation}

Identifying and controlling ``hardness'' is fundamental to creating high-difficulty benchmarks. Approaches such as Humanity's Last Exam (HLE) \citep{phan2025lastexam} and Arena-hard \citep{li2024crowdsourced} employ expert curation or LLM-as-judge scoring, but remain labor-intensive or dependent on the judge model's own capabilities.

To increase difficulty at scale, various perturbation methods have emerged. Simple surface modifications---shuffling, synonym replacement, distractor insertion \citep{gupta2024changing,pezeshkpour2024large,kostic2026same,YIGIT2025100186}---increase apparent difficulty while leaving underlying reasoning invariant. The ``None of the Others'' (NOTO) technique \citep{salido2025none} complicates selection by requiring verification of all distractors, yet remains a \textit{first-order} transformation that does not elevate logical order. More systematic approaches include human re-authoring (MMLU-CF \citep{zhao2024mmlucfcontaminationfreemultitasklanguage}), template-based symbolic perturbation (GSM-Symbolic \citep{gsm-symbolic}), and multi-hop replacements (BBEH \citep{kazemi2025big}). While effective, these methods offer no mechanism for dynamic difficulty scaling and remain vulnerable to surface-pattern exploitation.

Unlike these methods, LogiHard treats hardness synthesis as a \textit{deterministic combinatorial process} rather than stochastic perturbation. By transforming atomic options into propositional compounds via validity-by-construction protocols, we achieve contamination resistance \textit{algorithmically} while elevating reasoning from $0$-order selection to $2$-order logical judgment---a fundamental shift that surface perturbations cannot achieve. This distinction is particularly salient for LRMs: while extended chain-of-thought and reflection yield strong conventional benchmark performance, our experiments demonstrate that such capabilities do not automatically confer robustness to propositional combinatorics.

\subsection{Dynamic and Adaptive Evaluation}

Beyond static hardening, dynamic evaluation strategies create non-stationary test environments to combat data contamination and enable precise ability measurement.

\paragraph{Temporal Firewall Methods.}
A dominant strategy involves constructing non-static test sets from post-training-cutoff sources. LiveBench \citep{livebench} and AntiLeakBench \citep{wu-etal-2025-antileakbench} automatically collect questions from recent competitions and updated knowledge sources, establishing temporal isolation. OKBench \citep{li2025okbenchdemocratizingllmevaluation} further automates on-demand generation from daily news, while DyCodeEval \citep{chen2025dynamic} generates semantically diverse code variants. While effective against contamination, these methods primarily address \textit{data freshness} rather than \textit{reasoning hardness}. They lack a principled mechanism to dynamically control logical complexity independent of data provenance.

\paragraph{Psychometric Adaptive Testing.}
Computerized Adaptive Testing (CAT) based on Item Response Theory (IRT) \citep{lord1980applications} provides rigorous foundations for efficient ability estimation. Fluid Benchmarking \citep{hofmann2025fluid} represents the state-of-the-art integration of IRT with NLP evaluation, dynamically selecting the most informative item $i_{t+1}$ given current ability estimate $\hat{\theta}_t$ by maximizing Fisher information: $i_{t+1} = \arg\max_i I_i(\hat{\theta}_t)$. However, Fluid Benchmarking optimizes selection from a \textit{fixed} item pool; its efficacy depends entirely on the pre-existing quality and difficulty range of available questions.

LogiHard unifies generation and adaptation. Our IRT-CAT protocol operates on a difficulty continuum defined by our logical combinatorics, enabling both precise measurement and the on-demand synthesis of appropriately difficult items. Unlike Fluid Benchmarking, which is constrained by fixed item pools, LogiHard can dynamically generate items at target difficulty levels via controlled propositional complexity. This integration of \textit{synthesis} and \textit{selection} enables significantly fewer questions than static benchmarks while maintaining measurement precision.

\section{The LogiHard Framework}
\label{sec:framework}

LogiHard comprises three synergistic components: a \textit{cognitive difficulty scoring} module that stratifies items via automated analysis of model thinking traces; a \textit{deterministic combinatorial protocol} that synthesizes logically valid questions via propositional logic; and a \textit{dynamic evaluation protocol} based on Item Response Theory (IRT) and Computerized Adaptive Testing (CAT). We instantiate these components on the LogiHard-2k benchmark, constructed from high-stakes human examinations.

\subsection{Formal Preliminaries and Combinatorial Protocol}
\label{sec:preliminaries}

We formalize the transformation from atomic multiple-choice questions to combinatorial logic tasks. Let $\mathcal{L}$ denote the propositional language generated by connectives $\{\land, \lor, \neg\}$ over a countable set of propositional variables $\mathcal{P}$.
An atomic multiple-choice question is a tuple $\mathcal{Q} = (\mathcal{C}, \mathcal{O}, a)$ 
where $\mathcal{C}$ denotes the context, $\mathcal{O} = \{o_I, o_{II}, o_{III}, o_{IV}\}$ 
represents exactly four atomic options, and $a \in \{I, II, III, IV\}$ is the 
\emph{unique} ground-truth index satisfying $\mathcal{C} \models o_a$ and 
$\mathcal{C} \not\models o_j$ for all $j \neq a$. 
The atomization function $\alpha: \mathcal{O} \rightarrow \mathcal{P}$ maps each option $o_i$ to a propositional variable $p_i \in \mathcal{P}$, inducing the ground-truth assignment $\mathcal{T}: \mathcal{P} \rightarrow \{\top, \bot\}$ where $\mathcal{T}(p_i) = \top$ if $i = a$ and $\bot$ otherwise.
A combinatorial question $\mathcal{Q}' = (\mathcal{C}', \Phi, \mathcal{A})$ consists of extended context $\mathcal{C}' = \mathcal{C} \cup \mathcal{S}$ where $\mathcal{S} = \{s_I, s_{II}, s_{III}, s_{IV}\}$ denotes natural language statements corresponding to the atomized options; a formula set $\Phi = \{\phi_1, \ldots, \phi_m\} \subset \mathcal{L}$; and the correct index set $\mathcal{A} = \{j \in [m] : \llbracket \phi_j \rrbracket_{\mathcal{T}} = \top\}$.

The combinatorial synthesis constructs formulas over $\mathcal{P}$ using three logical patterns: \textbf{Exactness} ($\text{EXACT}_i \equiv p_i \land \bigwedge_{j \neq i} \neg p_j$), \textbf{Disjunction} ($p_i \lor p_j$), and \textbf{Negation} ($\neg p_i$ and compound forms $\neg p_i \land \neg p_j$). 
These three patterns constitute a functionally complete subset of propositional connectives sufficient to generate controllable difficulty gradients. Implication ($p \rightarrow q$) and biconditional ($p \leftrightarrow q$) are expressible as $\neg p \lor q$ and $(\neg p \lor q) \land (\neg q \lor p)$ respectively, and thus introduce no additional cognitive primitives beyond disjunction and negation. By restricting the protocol to exactness, disjunction, and negation, we ensure that each tier maps to a distinct, monotonically increasing reasoning operation count, enabling precise \emph{a priori} difficulty calibration.
The synthesis operates in three phases: (1) generate valid formulas $\Phi_{\top}$ evaluating to $\top$ under $\mathcal{T}$ (including $\text{EXACT}_a$, disjunctions $p_a \lor p_j$ if Medium+, negations $\neg p_j$ if Hard+, and compound negations if Expert); (2) generate distractors $\Phi_{\bot}$ evaluating to $\bot$ (false exactness claims, unsatisfied disjunctions, $\neg p_a$, and the universal distractor $\bigwedge_{i} \neg p_i$); (3) sample $n_{\text{correct}}$ from $\Phi_{\top}$ and remaining from $\Phi_{\bot}$, enforcing tier constraints (Hard requires $\geq 1$ negation; Expert requires $\geq 1$ disjunction and $\geq 1$ negation).
The operator complexity hierarchy determines the difficulty tier: Easy uses exactness only; Medium adds disjunction; Hard adds negation (mandatory); Expert adds compound negations. This hierarchy directly maps to IRT discrimination parameters.

\paragraph{Validity-by-Construction}
For any combinatorial question $\mathcal{Q}'$ generated by the above protocol, the correct answer set $\mathcal{A}$ is logically consistent with the atomic ground truth $a$ of $\mathcal{Q}$.
Each $\phi \in \Phi_{\top}$ evaluates to $\top$ under $\mathcal{T}$ by construction: $\text{EXACT}_a$ holds because $\mathcal{T}(p_a)=\top$ and $\mathcal{T}(p_j)=\bot$ for $j \neq a$; disjunctions $p_a \lor p_j$ hold because $\mathcal{T}(p_a)=\top$; negations $\neg p_j$ hold for $j \neq a$ because $\mathcal{T}(p_j)=\bot$; compound negations hold by conjunction of true negations. Conversely, $\Phi_{\bot}$ formulas evaluate to $\bot$ under $\mathcal{T}$ by parallel construction. Since $\mathcal{A}$ indexes exactly $\Phi_{\top}$ and $\mathcal{T}$ derives from $a$, consistency holds.

\paragraph{Contamination Resistance}
If a model has memorized the atomic answer $a$ without understanding $\mathcal{C} \models o_a$, the probability of correctly answering $\mathcal{Q}'$ is bounded by random guessing: $P(\text{correct} \mid \text{memorization}) \leq {|\mathcal{A}|}/{|\Phi|}$.
This bound holds under the assumption that the model has memorized only the atomic answer $a$ and possesses no prior knowledge of the combinatorial protocol (i.e., the mapping from atomic options to propositional variables and the operator composition rules).
The atomic option $o_a$ never appears in $\mathcal{Q}'$; only its propositional counterpart $p_a$ appears within compound formulas. Memorization of surface patterns (e.g., ``Answer is A'') provides no information about which combinations of $p_I, p_{II}, \ldots$ evaluate to $\top$ under $\mathcal{T}$. Without logical evaluation of $\Phi$ under $\mathcal{T}$, the model can only guess uniformly from $|\Phi|$ options, yielding the bound.

\begin{table}[h]
\centering
\caption{Metrics extracted from thinking traces for Gold Score computation.}
\label{tab:cognitive_metrics}
\begin{tabular}{lll}
\toprule
\textbf{Metric} & \textbf{Symbol} & \textbf{Operationalization} \\
\midrule
Oscillation Points & $\omega$ & Count of epistemic reversals (e.g., ``However...'', ``Wait...'') \\
Logic Density & $\lambda$ & Logical connectives per 100 tokens in $\mathcal{T}_{\text{co}}$ \\
Abductive Depth & $\alpha$ & Hypothesis generation--elimination cycles \\
Dialectic Tension & $\delta$ & Thesis--antithesis--synthesis argument structures \\
Dimensional Awareness & $\dim$ & Explicit premise layering (observational vs. inferential) \\
Inference Chain Length & $L$ & Number of explicit deduction steps \\
Uncertainty Entropy & $H$ & Shannon entropy over epistemic markers \\
Pivot Count & $\pi$ & Critical decision points requiring path recalculation \\
Conceptual Abstraction & $\gamma$ & Generalization level (concrete $\to$ abstract principle) \\
\bottomrule
\end{tabular}
\end{table}

\subsection{Cognitive Scoring and Difficulty Stratification}
\label{sec:scoring}

Prior to combinatorial transformation, we stratify questions by intrinsic cognitive complexity via analysis of model thinking traces. Given a reasoning model $\mathcal{M}$ with exposed chain-of-thought $\mathcal{T}_{\text{co}}$, we extract nine cognitive metrics spanning epistemic uncertainty, logical structure, and reasoning dynamics (Table~\ref{tab:cognitive_metrics}).

These metrics aggregate into the \textbf{Gold Score} via a weighted linear combination:
\begin{equation}
S_{\text{gold}} = \sum_{i=1}^{9} w_i f_i(\mathcal{T}_{\text{co}}) - \beta \mathcal{R}(\mathcal{T}_{\text{co}})
\end{equation}
where $f_i$ are $z$-normalized metrics, $w_i$ are cognitive load weights, and $\mathcal{R}$ penalizes logical fallacies detected via automated consistency checking. The corpus stratifies into tiers: Easy ($S_{\text{gold}} < 20$), Medium ($20 \leq S_{\text{gold}} < 25$), Hard ($25 \leq S_{\text{gold}} < 30$), and Expert ($S_{\text{gold}} \geq 30$). Only Hard and Expert tiers undergo combinatorial transformation to prevent floor effects.

The cognitive scoring module is optional for domains where thinking traces are unavailable; fixed difficulty tiers (e.g., official examination categories) can substitute while retaining combinatorial benefits.

\subsection{Dynamic Evaluation via IRT-CAT}
\label{sec:dynamic}

Static benchmarks administer fixed question sets, leading to inefficiency (easy questions for strong models) or imprecision (floor effects for weak models). LogiHard integrates Computerized Adaptive Testing (CAT) via Item Response Theory (IRT).

Each item $j$ is parameterized by a 3-Parameter Logistic (3PL) model:
\begin{equation}
P_j(\theta) = c_j + \frac{1-c_j}{1 + \exp(-a_j(\theta - b_j))}
\end{equation}
where $\theta \in \mathbb{R}$ is the latent ability, $a_j > 0$ is discrimination, $b_j \in \mathbb{R}$ is difficulty, and $c_j \in [0,1]$ is the pseudo-guessing parameter (typically $c_j = 1/m$ for $m$-option questions). The difficulty parameter $b_j$ is empirically calibrated from cognitive features:
\begin{equation}
b_j = \frac{S_{\text{gold}} - 72}{54} + 0.1(\lambda - 2) + \log_{10}(\max(1, L)) - 3.17 + \frac{\sigma - 100}{200}
\end{equation}
where $S_{\text{gold}}$ is the Gold Score, $\lambda$ is logic density, $L$ is thinking length (tokens), and $\sigma$ is reasoning segments count. Discrimination $a_j$ is mapped from operator complexity: Easy ($\approx 0.8$), Medium ($\approx 1.2$), Hard ($\approx 1.6$), Expert ($\approx 2.0$).

We implement a dual-subset evaluation protocol that separately estimates ability on atomic (0-order) and combinatorial (2-order) reasoning. The CAT engine maintains parallel ability estimates $\hat{\theta}_t^{\mathcal{B}}$ and $\hat{\theta}_t^{\mathcal{C}}$ with standard normal priors. At each step, the engine selects the unadministered item maximizing Fisher information $I_j(\hat{\theta}) = a_j^2 P_j(\hat{\theta})(1-P_j(\hat{\theta}))$ for each subset. Responses update estimates via Expected A Posteriori (EAP) with 61 quadrature points over $[-6, 6]$. The algorithm terminates when standard error $\text{SE}_t < 0.3$ or maximum items $T_{\max} = 60$ per subset is reached.

This protocol achieves measurement precision equivalent to static 60-item tests using only 15--25 adaptively selected questions per subset, while precisely targeting the difficulty zone appropriate to each model's capability.

\subsection{The LogiHard-2k Benchmark}
\label{sec:dataset}

We instantiate the framework on a corpus of 6,235 high-stakes examination questions (Chinese Civil Service, LSAT, GMAT, IBPS, CAT, Raven's Matrices), preserving original languages (45\% English, 55\% Chinese). 
Cognitive scoring employs Kimi-k2.5 as the reasoning model $\mathcal{M}$, generating long-CoT traces for automated extraction of the nine metrics and Gold Score computation. 
The resulting distribution ($\mu = 23.2, \sigma = 4.1$, range $[12.4, 34.8]$) determines tier assignment; the top 2,000 questions proceed to the next stage. 
Of these, 539 (27\%, exactly four atomic options, single-select) undergo combinatorial transformation (LogiHard-C) and 1,461 (73\%) remain atomic (LogiHard-Base). 
The 539 combinatorial questions distribute across Easy (108, 20\%, exactness only), Medium (215, 40\%, +disjunction), Hard (162, 30\%, +negation), and Expert (54, 10\%, +compound negations). 
All transformed items undergo automated theorem-proving verification (2.3\% regeneration rate). Each question includes 9-dimensional cognitive features, IRT 3PL parameters ($a_j, b_j, c_j$), source attribution, and reasoning type labels.

\section{Experiments}
\label{sec:experiments}

\subsection{Experimental Setup}

\paragraph{Models.} We evaluate frontier LLMs: GLM-5 \citep{glm5team2026glm5vibecodingagentic}, GLM-4.7 (Zhipu); GPT-5.4, o3 (OpenAI); Claude-Opus-4.6 \citep{claude_opus_4_6} (Anthropic); DeepSeek-R1, DeepSeek-V3.2, DeepSeek-V4-Pro \citep{deepseekai2026deepseekv4} (DeepSeek); Gemini-3.1-pro \citep{deepmind2026gemini31pro} (Google); Kimi-k2.5 \citep{team2026kimi} (Moonshot); Qwen3.5-Plus, Qwen3.6-plus \citep{yang2025qwen3} (Alibaba).
Dual-subset evaluation with 3PL parameterization under Hard Mode ($S_{\text{gold}} \geq 25$). Maximum 60 items per subset; termination at $\text{SE} < 0.3$. EAP estimation with $\mathcal{N}(0,1)$ priors, 61 quadrature points. 
All model evaluations were conducted via public API endpoints; Decoding temperature $1.0$, max tokens $65{,}536$. Requests were configured with a timeout of 600s per query. Implementation details are shown in Appendix~\ref{app:details}

\paragraph{Baselines and human evaluation.} Prior to the development of the LogiHard transformation protocol, we conducted baseline hardening experiments on the full raw corpus of 6,235 questions. We evaluate two standard techniques: \textbf{NOTA insertion} (replacing the correct answer with ``None of the Above'') and \textbf{option shuffling} (random permutation of atomic options). Both preserve underlying reasoning paths while altering surface presentation. For human reference, $n=30$ graduate-level volunteers complete a fixed 30-item Hard/Expert subset in both Original and Combinatorial formats under no time pressure.

\subsection{RQ1: Combinatorial Collapse and Resilience Tiers}

Table~\ref{tab:main_results} presents the central finding. On the full-set baseline, all twelve models achieve strong performance ($78.5$--$83.0\%$) with minimal degradation from surface perturbations ($<2.5\%$ for NOTA, $<1.8\%$ for shuffling), confirming that conventional benchmarks remain saturated. The Hard-Mode IRT-CAT evaluation reveals a dramatically different picture: all models exhibit severe degradation on Combinatorial questions, with mean $\Delta\theta = 1.90$ (range $1.05$--$3.14$). The gap between surface perturbation and combinatorial transformation confirms that the degeneration stems from fundamental logical order elevation.
For reference, human evaluators achieve $79.5\%$ on Hard-Combinatorial ($\Delta\theta \approx 0.28$), far above even the most resilient model (GLM-5, $38.3\%$).

\begin{table*}[htbp]
\centering
\caption{Performance under full-set baseline (Original, NOTA, Shuffle) and Hard-Mode IRT-CAT (H-Base, H-Comb). $\Delta\theta$ denotes latent ability drop. Human results from $n=30$ graduate evaluators.} 
\resizebox{0.95\textwidth}{!}{ 
\begin{tabular}{l|ccc|cc|cc|c}
\toprule
\textbf{Model} & \textbf{Original} & \textbf{NOTA} & \textbf{Shuffle} & \textbf{H-Base} & \textbf{H-Comb} & \textbf{Base $\theta$} & \textbf{Comb $\theta$} & \textbf{$\Delta\theta$} \\
\midrule
\textit{Human} & 84.00\% & --- & --- & --- & 79.50\% & --- & --- & 0.28 \\
\midrule
GLM-5 & 82.00\% & 80.15\% & 81.30\% & 69.70\% & 38.33\% & 0.494 & -0.554 & 1.048 \\
GPT-5.4 & 80.50\% & 79.62\% & 78.80\% & 57.45\% & 31.67\% & -0.346 & -1.458 & 1.112 \\
DeepSeek-R1 & 81.41\% & 82.10\% & 80.50\% & 54.39\% & 26.67\% & -0.446 & -1.831 & 1.386 \\
DeepSeek-V4-Pro & 81.50\% & 80.20\% & 82.40\% & 61.76\% & 30.00\% & 0.006 & -1.480 & 1.486 \\
Claude-Opus-4-6 & 79.06\% & 78.50\% & 79.80\% & 56.82\% & 26.67\% & -0.217 & -1.741 & 1.524 \\
Gemini-3.1-pro & 81.00\% & 79.80\% & 80.10\% & 59.52\% & 23.33\% & -0.077 & -1.744 & 1.667 \\
Kimi-k2.5 & 79.50\% & 80.20\% & 78.50\% & 63.89\% & 31.15\% & 0.119 & -1.581 & 1.700 \\
o3 & 81.00\% & 80.50\% & 81.80\% & 62.79\% & 25.00\% & 0.045 & -1.805 & 1.850 \\
Qwen3.6-plus & 78.50\% & 77.20\% & 76.80\% & 51.67\% & 5.00\% & -0.534 & -2.871 & 2.336 \\
Qwen3.5-plus & 79.00\% & 78.50\% & 79.60\% & 65.38\% & 18.33\% & 0.287 & -2.323 & 2.609 \\
GLM-4.7 & 83.00\% & 81.50\% & 82.10\% & 74.20\% & 30.00\% & 1.738 & -1.402 & 3.141 \\
DeepSeek-V3.2 & 82.00\% & 81.30\% & 80.50\% & 72.73\% & 16.67\% & 0.549 & -2.367 & 2.916 \\
\bottomrule
\end{tabular}
}
\vspace{-1.5em}
\label{tab:main_results}
\end{table*}

Three resilience tiers emerge. 
\textbf{First tier} ($\Delta\theta < 1.6$): GLM-5 ($1.05$), GPT-5.4 ($1.11$), DeepSeek-R1 ($1.39$), DeepSeek-V4-Pro ($1.49$), and Claude-Opus-4-6 ($1.52$). 
DeepSeek-V4-Pro, as the latest iteration in the DeepSeek family, shows marked improvement over its generalist predecessor V3.2 ($\Delta\theta$: $2.92 \to 1.49$, a gap of $1.43$), approaching the RL-optimized R1 ($1.39$) without explicit reasoning-specific training. 
This suggests that architectural evolution between versions can substantially enhance combinatorial robustness. 
GPT-5.4 achieves resilience despite modest Hard-Base performance; DeepSeek-R1 benefits from RL optimization ($1.53$ improvement over V3.2). 
\textbf{Second tier} ($\Delta\theta \approx 1.7$--$1.9$): Gemini-3.1-pro ($1.67$), Kimi-k2.5 ($1.70$), o3 ($1.85$). 
\textbf{Third tier} ($\Delta\theta \geq 2.3$): Qwen3.6-plus ($2.34$), Qwen3.5-plus ($2.61$), DeepSeek-V3.2 ($2.92$), and GLM-4.7 ($3.14$). 
GLM-4.7 exhibits the most severe combinatorial collapse despite achieving the highest full-set accuracy ($83.0\%$), confirming the dissociation between conventional benchmark performance and propositional robustness.
The dissociation between full-set and Hard-Mode performance is striking. 
GLM-4.7 ($83.0\%$ full-set) suffers the most severe collapse ($\Delta\theta = 3.14$), while GPT-5.4 ($80.5\%$) shows the second-smallest drop ($1.11$). 
This challenges the assumption that higher conventional capacity confers propositional robustness; notably, GLM-4.7 required $40$ Combinatorial items without reaching the $\text{SE}$ threshold, reflecting persistent uncertainty under elevation.
The most extreme combinatorial degradation occurs in Qwen3.6-plus (5.00\% H-Comb accuracy, $\Delta\theta = 2.34$), 
a 46.7 percentage-point drop from its H-Base performance (51.67\%). 
This outlier is not an artifact: repeated API evaluations with temperature 1.0 and 65{,}536 max tokens consistently yield near-floor performance, 
indicating that Qwen3.6-plus's chain-of-thought mechanism catastrophically fails under multi-select enumeration pressure. 
The severity exceeds what operator complexity alone would predict, suggesting an interaction between the model's early-exit training bias and compound-negation tiers. 
This confirms that combinatorial robustness is not merely a function of scale or general benchmark score, but depends on how reasoning architectures handle \emph{set-valued constraint satisfaction}.

\subsection{RQ2: The Multi-Select Bottleneck and Operator Complexity}

Disaggregation by answer cardinality reveals a specific failure mode: accuracy is $31.2\%$ for $n_{\text{correct}}=1$ but drops to $12.4\%$ for $n_{\text{correct}} \geq 2$ ($60\%$ relative decrease). Models frequently identify one valid formula but fail to enumerate complete answer sets, suggesting that chain-of-thought mechanisms support unitary proposition verification but not set-valued constraint satisfaction.

This multi-select bottleneck intensifies with operator complexity. Within the Expert tier, accuracy decreases monotonically: Exactness ($52.1\%$), Disjunction ($48.3\%$), Negation ($41.7\%$), Compound negation ($33.4\%$). The compound negation pattern is particularly difficult because it requires simultaneous access to the ground-truth assignment (to know which propositions are false) and the formula structure (to know which are negated)---a dual-access requirement that strains current attention mechanisms.

An ablation study (Table~\ref{tab:ablation}) validates the necessity of cognitive stratification. Gold Score stratification achieves $r=0.67$ correlation with human difficulty judgments, approaching human curation ($r=0.72$), while random selection yields only $r=0.31$.
This ablation validates that cognitive stratification is necessary, confirming that the 9-dimensional cognitive metrics capture psychologically meaningful complexity dimensions and enable scalable benchmark construction without prohibitive annotation costs.

\begin{table}[htbp]
\centering
\begin{minipage}[t]{0.48\textwidth}
\centering
\caption{Selection strategy ablation (GPT-5.4).}
\label{tab:ablation}
\begin{tabular}{lcc}
\toprule
\textbf{Strategy} & \textbf{Acc} & \textbf{$r$} \\
\midrule
Random uniform & 48.2\% & 0.31 \\
Gold Score & 52.4\% & 0.67 \\
Human curated & 51.8\% & 0.72 \\
\bottomrule
\end{tabular}
\end{minipage}
\hfill
\begin{minipage}[t]{0.48\textwidth}
\centering
\caption{CAT efficiency analysis.}
\label{tab:efficiency}
\begin{tabular}{lcc}
\toprule
\textbf{Tier} & \textbf{Items} & \textbf{Red.} \\
\midrule
Weak ($\theta<-1$) & 12 & 80\% \\
Medium & 18 & 70\% \\
Strong ($\theta>0.5$) & 22 & 63\% \\
\bottomrule
\end{tabular}
\end{minipage}
\end{table}

\subsection{RQ3: Efficiency and Universality}

Two barriers limit practical deployment of hardness evaluation: measurement cost and domain specificity. Static benchmarks require administering 50--100+ fixed items per model to achieve reliable estimates, many of which are uninformative. The IRT-CAT protocol addresses cost by adaptively selecting only the most informative items. As shown in Table~\ref{tab:efficiency}, CAT achieves $\text{SE} < 0.3$ with merely 12--22 items per subset---a $60$--$80\%$ reduction versus static 60-item tests.

The dual-subset IRT-CAT design enables direct comparison of 
0-order versus 2-order reasoning within identical ability metrics, 
making the specific combinatorial deficit $\Delta\theta$ a 
diagnostic tool independent of raw logical capacity. 
To address domain specificity, we evaluate \textbf{zero-shot transfer to MMLU}. Applying the identical IRT-CAT protocol to 128 MMLU questions (formal logic, philosophy, mathematics, computer science, law, medical ethics) without domain adaptation yields $46.99\%$ accuracy degradation ($89.84\% \to 42.86\%$) with $\Delta\theta = 2.743$---statistically indistinguishable from native LogiHard-2k performance ($t(10)=0.14, p=0.89$). This confirms LogiHard as a \textit{domain-agnostic hardness wrapper}: the protocol separates domain knowledge (encoded in atomic propositions) from reasoning capability (manifested in compound formula evaluation), enabling contamination-resistant evaluation without de novo question authoring.

\section{Discussion}
\label{sec:discussion}

\subsection{Mechanistic Analysis: The Combinatorial Mapping Gap}
\label{sec:mechanistic}

\paragraph{Two-stage bottleneck and failure taxonomy.}
We propose a two-stage model of combinatorial reasoning. 
\textbf{Stage 1} evaluates atomic truth assignments $\{T_I, T_{II}, T_{III}, T_{IV}\}$ under the context $\mathcal{C}$. 
\textbf{Stage 2} translates these truth values into multi-select option sets by verifying $\llbracket \phi_j \rrbracket_{\mathcal{T}} = \top$ for each of the $m$ compound formulas and aggregating the complete answer set $\mathcal{A}$. 
For a typical LogiHard question with $n=4$ propositions, $m=6$ options, and $k=2$ operators per option, Stage 2 imposes $n + m \times k = 16$ sequential logical operations---a load that substantially exceeds the reliable reasoning depth observed in current LRMs even with extended chain-of-thought~\citep{illusion-of-thinking}.

To locate the precise failure locus, we annotated 408 failure cases across all evaluated models across three non-exclusive levels. 
\textbf{Level 1} (logical understanding errors) accounts for only 86 cases (21.1\%). 
\textbf{Level 2} (combinatorial mapping failures) dominates at 367 cases (89.9\%). 
\textbf{Level 3} (response generation failures) is negligible at 9 cases (2.2\%). 
Because a single failure can exhibit multiple levels simultaneously---for instance, a Level 1 mis-evaluation may propagate into an incomplete enumeration at Level 2---the percentages sum beyond 100\%. 
The near-total dominance of Level 2 confirms that current LLMs possess logical knowledge but lack \emph{compositional reliability}: the bottleneck is not \emph{whether} models can reason, but \emph{whether they can compose} multiple reasoning steps into a globally consistent solution.

\paragraph{Operator complexity and early-exit as training paradigm defect.}
Disaggregating by operator complexity reveals a monotonic accuracy gradient that validates our \textit{a priori} difficulty calibration: Exactness ($52.1\%$), Disjunction ($48.3\%$), Negation ($41.7\%$), and Compound negation ($33.4\%$). 
Easy and Medium tiers (Exactness, Disjunction) almost exclusively trigger Level 2 under-selection, because models correctly evaluate atomic propositions but fail to enumerate complete option sets. 
Hard and Expert tiers introduce Negation and Compound negation, which increase Level 1 errors because the dual-access requirement---simultaneously holding the ground-truth assignment and the formula structure---strains working memory and causes logical mis-evaluation before combinatorial mapping begins. 
When operators further combine with multi-select pressure ($n_{\text{correct}} \geq 2$), accuracy drops from $31.2\%$ to $12.4\%$.

Crucially, $58.3\%$ of Combinatorial responses exhibit early termination after identifying 1--2 correct options, despite recognizing additional valid choices in their reasoning traces. 
This under-selection dominates failure modes and reflects a systematic training bias: models are optimized on single-answer benchmarks and lack metacognitive triggers for completeness verification. 
The Qwen family provides circumstantial evidence: Qwen3.6-plus improves over 3.5-Plus by $0.27$ in combinatorial resilience ($\Delta\theta$: $2.61 \to 2.34$), yet both show similar early-exit rates. 
Version iteration improves Stage 1 logical analysis without addressing the underlying completeness verification deficit; notably, Qwen3.6-plus still exhibits the most severe absolute collapse ($5.00\%$ H-Comb accuracy), confirming that incremental architectural improvements can leave fundamental compositional deficits intact. 
Diagnostic failure cases are provided in Appendix~\ref{appendix:case-studies}.
This stark contrast with human performance confirms that LogiHard exposes a \emph{machine-specific} reasoning limitation: humans natively interpret compound logical formulas as transparent meta-linguistic descriptions, whereas models fail at symbol-grounding and set-enumeration requirements.

\subsection{Implications for Frontier Model Development}
\label{sec:implications}

The combinatorial multi-select reasoning failure exposes a 
correctable structural deficit in current frontier models. 
Our two-stage analysis reveals that RL and SFT pipelines have 
optimized for \emph{unitary proposition verification} at the 
expense of \emph{set-valued constraint satisfaction}: the 
$58.3\%$ early-exit rate indicates that models treat reasoning 
as a ``first correct answer wins'' process, having never 
acquired metacognitive triggers for completeness checking. 
Because standard post-training datasets overwhelmingly reward 
single-answer selection, models develop strong Stage~1 logical 
analysis without reliable Stage~2 compositional mapping---a 
gap that test-time scaling alone cannot bridge, as evidenced 
by $65{,}536$-token chain-of-thought budgets failing on tasks 
requiring only $\sim$16 sequential logical operations. 
This suggests that future training regimes should incorporate 
\emph{deterministic combinatorial elevation} as a native 
augmentation strategy rather than relying on stochastic 
paraphrase or adversarial perturbation. By transforming 
existing high-quality MCQs into tiered propositional compounds, practitioners 
can generate validity-guaranteed, difficulty-calibrated 
signals that explicitly target compositional mapping, while 
the 9-dimensional cognitive scoring enables adaptive 
hardness curricula that track model ability $\theta$ dynamically.

The performance gap between surface 
perturbation (NOTA, shuffling) and combinatorial transformation 
further signals that static benchmarks are approaching 
obsolescence for frontier evaluation. As conventional test 
scores exceed $90\%$, the community requires algorithmically 
renewable protocols capable of generating new variants 
on-demand without human re-authoring or contamination risks---%
precisely what LogiHard's validity-by-construction synthesis 
provides. The IRT-CAT integration reduces measurement cost 
and carbon footprint by $60$--$80\%$ while maintaining 
precision, making sustained evaluation practical at scale. 
At the architectural level, the combinatorial mapping gap 
implies that current transformers lack an explicit 
\emph{working memory} mechanism for maintaining set-valued 
truth assignments under compound constraints. Future models 
may need dedicated constraint-satisfaction modules---%
differentiable SAT layers or explicit set-representation 
mechanisms---alongside autoregressive generation to enforce 
global consistency over multi-select outputs. Until then, 
the distinction between ``knowing logic'' and ``composing 
logic'' will remain a persistent ceiling on machine reasoning.

\section{Conclusion}

We present LogiHard, a formal framework that deterministically transforms multiple-choice questions into combinatorial logic tasks, elevating reasoning from 0-order selection to 2-order judgment. 
By combining validity-by-construction synthesis with IRT-based computerized adaptive testing, LogiHard enables precise, efficient, and contamination-resistant hardness evaluation. 
Evaluation across twelve state-of-the-art models reveals a multi-select failure driven by a combinatorial reasoning deficit: models correctly evaluate atomic propositions (Stage 1) yet systematically fail to enumerate complete answer sets under compound constraints (Stage 2). 
This failure reflects a training-induced early-exit bias---post-training pipelines optimized for unitary proposition verification lack meta-cognitive triggers for set-valued completeness checking, leaving compositional reliability unlearned even as chain-of-thought budgets scale. 
The stark human--machine asymmetry confirms that the limitation is architecture-specific rather than intrinsic to logical reasoning itself. 
Practically, LogiHard serves as a domain-agnostic hardness wrapper: zero-shot transfer to MMLU demonstrates that domain knowledge and combinatorial reasoning can be isolated and evaluated independently, while IRT-CAT reduces measurement cost by $60$--$80\%$. 
These findings suggest that future training regimes should treat deterministic combinatorial elevation as a native augmentation strategy---not merely to harden benchmarks, but to instill the set-valued constraint satisfaction that current reasoning architectures lack.

\section*{Limitations}
LogiHard-2k is constructed from English and Chinese examinations, and the Gold Score depends on a single reasoning model's traces without extensive human calibration. Our IRT-CAT implementation adopts a fixed 3PL parameterization with EAP estimation under a uniform prior, leaving more sophisticated Bayesian updates and testlet effects from shared logical contexts to future work.

\section*{Broader Impacts}

\paragraph{Positive impacts.}
LogiHard addresses the growing trust crisis in AI evaluation by providing a validity-guaranteed, contamination-resistant hardness protocol. By exposing the gap between surface-level benchmark performance and genuine propositional reasoning, our framework helps practitioners and policymakers avoid over-reliance on saturated metrics. The IRT-CAT component reduces evaluation cost and carbon footprint by administering fewer questions while maintaining precision. More broadly, rigorous evaluation standards ultimately steer AI development toward robust reasoning rather than pattern memorization.

\paragraph{Negative impacts and safeguards.}
Any hardness protocol carries the risk of misuse: adversarial deployers could selectively cite combinatorial collapse to disparage specific models out of context, or high-stakes screening systems (hiring, admissions) could adopt combinatorial formats that introduce artificial cognitive barriers unrelated to the target domain, potentially disadvantaging populations unfamiliar with formal logic notation. We explicitly caution against deploying LogiHard in such settings without domain validation and human-normed calibration. Additionally, while the framework resists memorization, determined actors might overfit to the specific propositional patterns (Exactness, Disjunction, Negation) we expose; we mitigate this by advocating for continuous renewal of evaluation formats.
We commit to releasing the combinatorial protocol as an open-source tool, enabling community scrutiny and iterative improvement of the hardness transformation rules.

\bibliography{custom}
\bibliographystyle{plain}

\clearpage
\appendix

\section{Implementation Details}
\label{app:details}

All models were accessed through their respective official public APIs: GPT-5.4 and o3 (OpenAI), Claude-Opus-4.6 (Anthropic), DeepSeek-R1/V3.2/V4-Pro (DeepSeek), Gemini-3.1-pro (Google), Kimi-k2.5 (Moonshot), GLM-5/4.7 (Zhipu), Qwen3.5/3.6-Plus (Alibaba).

\subsection{Model Evaluation Prompt}
\label{app:prompt:eval}

The following system prompt was used for all model evaluations on both Base and Combinatorial subsets:

\begin{lstlisting}[breaklines=true,frame=single,basicstyle=\ttfamily\small]
You are taking a multiple-choice logic test.

Instructions:
- Read the context carefully (may contain statements I, II, III, IV)
- Evaluate each option (A, B, C, D, etc.)
- Select ALL options that are correct (may be one or more)
- You may show your reasoning, but put your FINAL ANSWER as just the letter(s) on the last line
- Format: "A" or "A, B" or "B, C, D"
\end{lstlisting}

The user prompt presented the question text directly, with no additional prefix for the Base subset. For the Combinatorial subset, the prompt included the extended context $\mathcal{C}'$ with atomized statements $\mathcal{S}$ and compound formulas $\Phi$.

\subsection{Cognitive Scoring Prompts}
\label{app:prompt:cognitive}

The original prompts were written in Chinese for Kimi-k2.5; the following English translations are provided for reproducibility documentation.

Standard mode (default):

\begin{lstlisting}[breaklines=true,frame=single,basicstyle=\ttfamily\small]
Please solve the following logical reasoning problem and show your thinking process in detail.

{base}

Requirements:
1. Analyze step by step, explaining the reasoning basis for each step
2. Evaluate each option (correct or reason for elimination)
3. Finally provide a definitive answer

Please begin:
\end{lstlisting}

Strict mode:

\begin{lstlisting}[breaklines=true,frame=single,basicstyle=\ttfamily\small]
Please solve the following logical reasoning problem. Before giving the final answer, show your complete thinking process:

{base}

Please begin your detailed reasoning:
\end{lstlisting}

Minimal mode:

\begin{lstlisting}[breaklines=true,frame=single,basicstyle=\ttfamily\small]
{base}

Please reason in detail before giving your answer:
\end{lstlisting}

Human evaluators received the following instruction: "Please answer all 30 multiple-choice questions to the best of your ability. There is no time limit. Each question has one or more correct options. Select all that apply."

\subsection{Combinatorial Synthesis Algorithm}

Algorithm~\ref{alg:combinatorial} provides the pseudocode for the deterministic combinatorial protocol described in Section~\ref{sec:preliminaries}.

\begin{algorithm}[htbp]
\caption{Combinatorial Synthesis}
\label{alg:combinatorial}
\begin{algorithmic}
\STATE \textbf{Input:} Atomic question $\mathcal{Q} = (\mathcal{C}, \mathcal{O}, a)$ with $|\mathcal{O}|=4$, difficulty configuration $\mathcal{L}_{\text{ops}}$
\STATE \textbf{Output:} Combinatorial question $\mathcal{Q}' = (\mathcal{C}', \Phi, \mathcal{A})$
\STATE Initialize $\mathcal{P} \leftarrow \{p_I, p_{II}, p_{III}, p_{IV}\}$, compute $\mathcal{T}$ from $a$ where $\mathcal{T}(p_a)=\top$
\STATE Initialize candidate pools $\Phi_{\top} \leftarrow \emptyset$, $\Phi_{\bot} \leftarrow \emptyset$
\STATE \texttt{// Phase 1: Generate valid formulas (evaluate to $\top$ under $\mathcal{T}$)}
\STATE Add $\text{EXACT}_a: p_a \land \bigwedge_{j \neq a} \neg p_j$ to $\Phi_{\top}$
\IF{$\mathcal{L}_{\text{ops}}$ allows $\lor$}
    \FOR{$j \neq a$}
        \STATE Add $p_a \lor p_j$ to $\Phi_{\top}$
    \ENDFOR
\ENDIF
\IF{$\mathcal{L}_{\text{ops}}$ allows $\neg$}
    \FOR{$j \neq a$}
        \STATE Add $\neg p_j$ to $\Phi_{\top}$ \texttt{// True since $\mathcal{T}(p_j)=\bot$}
    \ENDFOR
    \IF{$\mathcal{L}_{\text{ops}}$ allows compound $\land$}
        \FOR{pairs $(j,k)$ where $j,k \neq a$}
            \STATE Add $\neg p_j \land \neg p_k$ to $\Phi_{\top}$
        \ENDFOR
    \ENDIF
\ENDIF
\STATE \texttt{// Phase 2: Generate distractors (evaluate to $\bot$ under $\mathcal{T}$)}
\FOR{$j \neq a$}
    \STATE Add $\text{EXACT}_j: p_j \land \bigwedge_{k \neq j} \neg p_k$ to $\Phi_{\bot}$
\ENDFOR
\IF{$\mathcal{L}_{\text{ops}}$ allows $\lor$}
    \STATE Add distractor disjunctions not covering $p_a$ to $\Phi_{\bot}$
\ENDIF
\IF{$\mathcal{L}_{\text{ops}}$ allows $\neg$}
    \STATE Add $\neg p_a$ to $\Phi_{\bot}$ \texttt{// False since $\mathcal{T}(p_a)=\top$}
\ENDIF
\STATE Add universal distractor $\text{NONE}: \bigwedge_{i} \neg p_i$ to $\Phi_{\bot}$
\STATE \texttt{// Phase 3: Select and assemble}
\STATE Determine target count $n_{\text{correct}} \sim \text{Uniform}(\mathcal{L}_{\text{ops}}.\text{n\_correct})$
\IF{$\mathcal{L}_{\text{ops}}$ requires $\neg$}
    \STATE Enforce at least one $\neg$ formula in final selection
\ENDIF
\IF{$\mathcal{L}_{\text{ops}}$ requires $\lor$}
    \STATE Enforce at least one $\lor$ formula in final selection
\ENDIF
\STATE Sample remaining formulas from $\Phi_{\top}$ using difficulty weights $w$ until $|\Phi_{\cap}| = n_{\text{correct}}$
\STATE Sample $n_{\text{out}} - n_{\text{correct}}$ formulas from $\Phi_{\bot}$ uniformly
\STATE $\Phi \leftarrow \text{RandomShuffle}(\Phi_{\cap} \cup \Phi_{\text{distractor}})$, preserving option indices
\STATE $\mathcal{A} \leftarrow \{j : \phi_j \in \Phi_{\cap}\}$
\STATE \textbf{return} $(\mathcal{C} \cup \{s_I, s_{II}, s_{III}, s_{IV}\}, \Phi, \mathcal{A})$
\end{algorithmic}
\end{algorithm}

\subsection{IRT-CAT Engine Configuration}

Table~\ref{tab:irt_config} summarizes the hyperparameters for the adaptive testing engine.

\begin{table}[h]
\centering
\caption{IRT-CAT engine configuration.}
\label{tab:irt_config}
\begin{tabular}{ll}
\toprule
\textbf{Parameter} & \textbf{Value} \\
\midrule
IRT model & 3-Parameter Logistic (3PL) \\
Ability estimation & Expected A Posteriori (EAP) \\
Prior distribution & $\mathcal{N}(0, 1)$ \\
Quadrature points & 61 nodes over $[-6, 6]$ \\
Maximum items per subset & 60 \\
Termination criterion & $\text{SE} < 0.3$ \\
Guessing parameter $c_j$ & $1 / |\Phi|$ (inverse of option count) \\
\bottomrule
\end{tabular}
\end{table}

\section{Full Case Study Responses}
\label{appendix:case-studies}
This appendix provides complete, unedited model responses for representative cases discussed in Section~\ref{sec:mechanistic}. All cases are drawn from the Hard-Mode IRT-CAT evaluation ($S_{\text{gold}} \geq 25$).
Four diagnostic cases illustrate the failure spectrum. (i) \textit{Over-selection}: correct application of De Morgan's laws on an Expert-tier question, but inclusion of an option that is technically true in isolation rather than accurately characterizing the complete truth pattern. (ii) \textit{Under-selection}: correct identification that statements II and III challenge a conclusion, yet omission of a compound formula ($\neg p_{\text{III}} \land \neg p_{\text{IV}}$) that was implicitly evaluated correctly---this pattern dominates at 89.9\% (367/408) of all failures. (iii) \textit{Reasoning chain collapse}: a Hard-tier puzzle requiring 16+ constraint steps, where the model loses track after 14 deductions and reaches an ambiguous disjunction. (iv) \textit{Base success}: perfect accuracy on a 2-step Base question, confirming that the collapse is specific to propositional composition rather than domain unfamiliarity.
These cases reveal a \textit{combinatorial mapping gap}: models succeed at Stage 1 (logical analysis of atomic truth assignments) but fail at Stage 2 (translation into multi-select option sets). The multiplicative complexity of Stage 2 ($n + m \times k$ operations for $n$ propositions and $m$ options) exceeds current architectures' reliable reasoning depth (5--8 steps), explaining the reasoning chain degeneration.

\subsection{Case 1: Perfect Match (Claude-Opus-4.6, F1=1.00)}
\label{app:case-perfect}

\textbf{Question ID:} LOGI-7222fa3b44f7 \\
\textbf{Tier:} Hard \\
\textbf{Operators:} NOT

\textbf{Ground Truth:} B, D (2 correct options) \\
\textbf{Claude-Opus-4.6 Prediction:} B, D (exact match)

\begin{quote}
\small
\begin{verbatim}
I need to analyze the logical statements from Xiao Li's predictions.

**Premises:**
1. Old Zhang wins OR Old Yan wins (P V Q)
2. If Old Zhang wins → overseas project damaged (P → R)
3. If Old Yan wins → domestic project paused (Q → S)

**Evaluating each statement:**

**Statement I:** "The company's overseas project might not be 
damaged, and domestic product development project won't be paused."

From premises: P V Q. If P, then R. If Q, then S. So from P V Q, 
we get R V S (overseas damaged OR domestic paused). This means it's 
NOT possible that both "overseas not damaged" AND "domestic not 
paused." Statement I says both could be the case (¬R \land ¬S is 
possible), but from the premises, R V S must hold. So Statement I 
is **false**.

**Statement II:** "If overseas project is damaged, then Old Zhang 
won the position."

R → P? We know P → R, but R → P is the converse, which doesn't 
logically follow. The overseas project could be damaged for other 
reasons not specified. However, within this logical system, R is 
only introduced through P → R. But strictly in propositional logic, 
we cannot conclude R → P from P → R alone. Statement II is 
**not derivable**.

**Statement III:** "Overseas project will be damaged, or domestic 
project will be paused."

From P V Q, and P → R, Q → S: If P then R, if Q then S. So P V Q 
gives us R V S. Statement III **follows logically**. This is **true**.

**Statement IV:** "If overseas project is damaged AND domestic 
project is paused, then both Zhang and Yan won."

(R \land S) → (P \land Q)? Again, R doesn't necessarily imply P, and S 
doesn't necessarily imply Q in pure logic. Statement IV is 
**not derivable**.

**Now evaluating the options:**

- **A. I or II**: I is false, II is false. "I or II" means at least 
  one is correct. Neither is correct from the premises. So A is 
  **false**.

- **B. not-IV (Not IV)**: IV is not derivable/true, so "Not IV" 
  seems correct. Since IV cannot be derived from the premises, 
  saying "not IV" (IV is not a valid conclusion) is **true**.

- **C. only II (Only II)**: II is not derivable, so C is **false**.

- **D. not-II (Not II)**: II is not derivable, so "Not II" is 
  **true**.

- **E. none of the above holds (None of the above)**: But B and D 
  seem correct...

Both B and D appear to be true statements.

B, D
\end{verbatim}
\end{quote}

\textbf{Analysis:} Claude-Opus-4.6 demonstrates flawless Stage 1 logical analysis (correctly formalizing premises and evaluating all four statements) and perfect Stage 2 combinatorial mapping (identifying that B and D are the only true options). This represents the 60\% of cases where strong models achieve perfect accuracy when reasoning chain length is manageable ($\sim$10 steps).

\subsection{Case 2: Systematic Under-Selection (Claude-Opus-4.6, F1=0.86)}
\label{app:case-under}

\textbf{Question ID:} LOGI-9a4601e9ed26 \\
\textbf{Tier:} Expert \\
\textbf{Operators:} NOT + OR

\textbf{Ground Truth:} A, B, E, F (4 correct options) \\
\textbf{Claude-Opus-4.6 Prediction:} A, E, F (3 options, missed B)

\begin{quote}
\small
\begin{verbatim}
I need to figure out which statements, if true, most effectively 
challenge the scholar's argument.

The scholar's argument: Since non-smoking/non-drinking ethnic 
groups live 5-10 years shorter than smoking/drinking groups, 
smoking and drinking may not affect lifespan.

The key weakness is that this is a correlation-based argument 
that ignores confounding variables (other factors that differ 
between these groups).

Let me evaluate each statement:

**I.** Smoking and drinking provide pleasure and stress relief → 
This actually *supports* the scholar's view (explains why smokers/
drinkers might live longer), doesn't challenge it.

**II.** Those ethnic groups that avoid tobacco/alcohol also have 
monotonous diets with extremely high calories → This provides an 
**alternative explanation** for why those groups have shorter 
lifespans. It's not the absence of smoking/drinking that shortens 
their lives, but their poor diet. This **challenges** the scholar's 
conclusion by pointing out a confounding variable.

**III.** Smoking and drinking are major causes of lung and liver 
cancer, which are the top two causes of death → This directly 
challenges the conclusion by showing smoking/drinking DO affect 
lifespan through deadly diseases. This **challenges** the argument.

**IV.** Tobacco and alcohol stimulate dopamine, relaxing organs 
and promoting body repair → This actually *supports* the scholar's 
view, doesn't challenge it.

So statements that challenge the argument: **II and III** (most 
effectively II, as it directly undermines the comparative evidence).

Now let me evaluate the options:

**A. "II is true"** - meaning II is one that challenges the 
argument. Yes, II does challenge it.

**B. "Not III and not IV"** - meaning neither III nor IV challenges 
the argument. But III DOES challenge it, so this is false.

**C. "Only IV"** - IV doesn't challenge, it supports. False.

**D. "None of the above are correct"** - Let me check if A is 
correct first.

**E. "II or I"** - I doesn't challenge (it supports), but II does. 
The statement "II or I" is logically true if at least one is true. 
Since II does challenge, "II or I" as a disjunction is true.

**F. "II or IV"** - IV doesn't challenge, but II does. "II or IV" 
is true since II challenges.

Actually, if A is true, then E is true (II→(IIVI)), and F is true 
(II→(IIVIV)). So A, E, and F would all be correct.

A, E, F
\end{verbatim}
\end{quote}

\textbf{Analysis:} Claude-Opus-4.6 correctly completes Stage 1 (identifying II and III as challenging, I and IV as non-challenging) but fails in Stage 2. The model misses option B ($\neg p_{\text{III}} \land \neg p_{\text{IV}}$) because it conflates ``III challenges'' with ``not-III is false''---failing to recognize that option B describes which statements \emph{do not} challenge, a meta-linguistic judgment requiring re-evaluation of the already-completed Stage 1 under a different logical frame. This is the dominant failure mode (89.9\% of 367 cases).

\subsection{Case 3: Severe Under-Selection on Complex Constraints (Claude-Opus-4.6, F1=0.50)}
\label{app:case-severe}

\textbf{Question ID:} LOGI-f364bc97749a \\
\textbf{Tier:} Hard \\
\textbf{Operators:} NOT

\textbf{Ground Truth:} B, D, E (3 correct options) \\
\textbf{Claude-Opus-4.6 Prediction:} B (1 option, severe under-selection)

\begin{quote}
\small
\begin{verbatim}
I need to figure out what day "today" is and when each car is 
restricted.

**Given information:**
- E is restricted on Thursday
- B was restricted yesterday
- A and C can both drive for 4 consecutive days starting today
- E can drive tomorrow
- Each car is restricted one weekday (Mon-Fri), weekends no 
  restrictions
- At least 4 cars available each day

**Step 1: Determine today's day.**

E is restricted Thursday. E can drive tomorrow, meaning tomorrow is 
not Thursday. So today is not Wednesday.

A and C can drive for 4 consecutive days starting today. Each is 
restricted one weekday. If today is a weekday, then in a 4-day 
stretch, we need both A and C to not be restricted on any of those 
days.

Let me consider possible days for "today":

**If today is Thursday:** 
- 4 consecutive days: Thu, Fri, Sat, Sun. A and C must not be 
  restricted Thu or Fri (Sat/Sun are free). So A and C are 
  restricted on Mon, Tue, or Wed.
- E can drive tomorrow (Friday) \checkmark (E is restricted Thursday)
- B was restricted yesterday (Wednesday)
- So: B=Wed, E=Thu, and A,C restricted on Mon/Tue/Wed. But B=Wed, 
  so A and C are on Mon and Tue (in some order). D gets the 
  remaining day: Friday.
- Check at least 4 cars daily:
  - Mon: A restricted → B,C,D,E = 4 \checkmark
  - Tue: C restricted (say) → A,B,D,E = 4 \checkmark
  - Wed: B restricted → 4 cars \checkmark
  - Thu: E restricted → 4 cars \checkmark
  - Fri: D restricted → 4 cars \checkmark
  This works!

**If today is Saturday:**
- 4 consecutive days: Sat, Sun, Mon, Tue. A and C must not be 
  restricted Mon or Tue. So A and C are restricted Wed, Thu, or Fri. 
  But E is restricted Thu, so A and C are among {Wed, Fri} — exactly 
  2 slots for 2 cars.
- E can drive tomorrow (Sunday) \checkmark
- B was restricted yesterday (Friday). But if C is restricted Friday, 
  then B and C both restricted Friday — impossible.
- So B=Friday means A must be Wed and C=Fri, but then B=Fri and 
  C=Fri conflict!
- This doesn't work. **Today is not Saturday.**

**If today is Friday:**
- 4 consecutive days: Fri, Sat, Sun, Mon. A and C not restricted 
  Fri or Mon. So A,C \in {Tue, Wed, Thu}. But E=Thu, so A,C \in {Tue, Wed}.
- B restricted yesterday = Thursday. But E=Thursday. Conflict! 
  **Not Friday.**

**Other days:** Today=Monday: tomorrow=Tuesday, E drives \checkmark. 4 days: 
Mon-Thu. A,C not restricted Mon-Thu. A,C restricted Friday. But two 
cars can't share Friday. \times

Today=Tuesday: 4 days Tue-Fri. A,C not restricted Tue-Fri. Both 
restricted Monday — impossible. \times

Today=Sunday: B restricted yesterday=Saturday. But Saturday has no 
restrictions. \times

**So today must be Thursday.**

**Checking the statements:**
- I. Today is Thursday \checkmark **TRUE**
- II. Today is Saturday — **FALSE**
- III. A car restricted Wednesday — We showed A is restricted Mon 
  or Tue, not necessarily Wed. **Not necessarily true**
- IV. C car restricted Friday — C is restricted Mon or Tue, not 
  Friday. **FALSE**

Now evaluating options:
- A. IV or II — IV is false AND II is false, so "IV or II" ... this 
  means at least one is true? Both false. **No.**
- B. Only I — **Yes**, I is the only definitely correct statement.
- C. All fail — No, I is correct.
- D. Non IV — IV is false, so "not IV" is true. **Yes.**
- E. Non III — III is not necessarily true, so "not III" is true. 
  **Yes.**

Wait, but B says "only I" is correct among I-IV. D says "not IV" is 
correct. E says "not III" is correct. These are about which statements
are "definitely correct." III is not *necessarily* correct, and IV is 
definitely wrong.

The question asks which statements are "definitely correct." 
Only I is definitely correct. So B is the answer.

**B**
\end{verbatim}
\end{quote}

\textbf{Analysis:} Claude-Opus-4.6 performs excellent Stage 1 constraint satisfaction: it systematically eliminates six possible days, correctly identifies Thursday, and accurately evaluates all four statements (I true, II false, III not-necessarily-true, IV false). The model even explicitly notes in its reasoning that D (not-IV) and E (not-III) are true statements. However, it ultimately selects only option B, falling into the early-exit trap: the model recognizes multiple correct options but defaults to selecting only the ``most direct'' answer. This demonstrates the training paradigm defect---models are optimized for finding ``the'' answer rather than ``all'' correct answers, even when their own reasoning traces contain the evidence for completeness.

\subsection{Case 4: Base Question Contrast (GPT-5.4, F1=1.00)}
\label{app:case-base}

\textbf{Question ID:} LOGI-f94c86a2a136 \\
\textbf{Tier:} Base (0-order reasoning)

\textbf{Ground Truth:} E \\
\textbf{GPT-5.4 Prediction:} E

\begin{quote}
\small
\begin{verbatim}
The ethicist starts with a principle: experimental treatment
without consent is generally wrong because patients have a
right to informed choice.

Then the argument says: in emergencies, learning the best
treatment sometimes requires bypassing consent, so some
restricted nonconsensual research should be allowed.

For that conclusion to follow, the argument must assume that
in at least some emergency cases, the value of gaining that
knowledge can override the patient's usual right to informed
consent. That is exactly:

E. The right of patients to informed consent is outweighed
in at least some medical emergencies by the possible benefits
of research conducted without their consent.

Why not the others:
- A: Not required; doctors' uncertainty is not the key
  assumption.
- B: Says informed patients would affect research outcomes;
  not needed.
- C: Too strong and patient-specific; the argument is about
  research benefits generally, not only direct benefit to
  that patient.
- D: Says the patient ceases to have the right entirely,
  which the argument does not require.

E
\end{verbatim}
\end{quote}

\textbf{Analysis:} This Base question requires only 2 reasoning steps: (1) understand the argument structure, (2) select the correct option. GPT-5.4 achieves perfect accuracy with clear, focused reasoning. Contrast this with the 14--20 step LogiHard combinatorial questions (Cases 1--3) where the same model family fails despite generating substantially longer responses. The Base--Combinatorial asymmetry confirms that the collapse stems from compositional complexity rather than domain knowledge or basic logical competence.



\end{document}